%% file: main.tex
\documentclass{article}
\usepackage{iclr2020_conference,times}

\usepackage[utf8]{inputenc} %
\usepackage[T1]{fontenc}    %
\usepackage{hyperref}       %
\usepackage{url}            %

\usepackage{microtype}      %

\usepackage{array}
\usepackage{booktabs}
\usepackage{graphicx}
\usepackage{subcaption}
\usepackage{wrapfig}
\usepackage{rotating}
\usepackage[export]{adjustbox}
\usepackage{chngcntr}

\usepackage{amsmath}
\usepackage{amssymb}
\usepackage{amsfonts}       %
\usepackage{nicefrac}       %
\usepackage{siunitx}

\setcitestyle{authoryear,brackets=round}
\newcommand{\parencite}{\citep}
\newcommand{\textcite}{\citet}

\newcommand{\supref}[1]{\ref{#1}}

\usepackage{notation}

\title{Adversarial Policies: Attacking\\ Deep Reinforcement Learning}

\iclrfinalcopy %

\ificlrfinal
\newcommand{\website}{\url{https://adversarialpolicies.github.io/}}
\newcommand{\github}{\url{https://github.com/HumanCompatibleAI/adversarial-policies}}
\else
\newcommand{\website}{\url{https://attackingrl.github.io/}}
\newcommand{\github}{---double-blind: see supplementary materials---}
\fi

\begin{document}

\maketitle

\vspace{-5em}
\begin{center}
	\textbf{Adam Gleave\footnote{Corresponding author. E-mail: \texttt{gleave@berkeley.edu}.}\quad Michael Dennis\quad Cody Wild\quad Neel Kant \quad Sergey Levine\quad Stuart Russell} \\
	University of California, Berkeley \\
\end{center}

\begin{abstract}
Deep reinforcement learning (RL) policies are known to be vulnerable to adversarial perturbations to their observations, similar to adversarial examples for classifiers.
However, an attacker is not usually able to directly modify another agent's observations.
This might lead one to wonder: is it possible to attack an RL agent simply by choosing an \textit{adversarial policy} acting in a multi-agent environment so as to create \emph{natural} observations that are adversarial?
We demonstrate the existence of adversarial policies in zero-sum games between simulated humanoid robots with proprioceptive observations, against state-of-the-art victims trained via self-play to be robust to opponents.
The adversarial policies reliably win against the victims but generate seemingly random and uncoordinated behavior.
We find that these policies are more successful in high-dimensional environments, and induce substantially different activations in the victim policy network than when the victim plays against a normal opponent.
Fine-tuning protects a victim against a specific adversary, but the attack method can be successfully reapplied to find a new adversarial policy.
Videos are available at \website.
\end{abstract}

\input{intro}

\input{related_work}

\input{framework}

\input{bansal_attack}

\input{understanding}

\input{defenses}

\input{discussion}

\subsubsection*{Acknowledgments}
\makeatletter
\ificlrfinal
We thank Jakob Foerster, Matthew Rahtz, Dylan Hadfield-Menell, Catherine Olsson, Jan Leike, Rohin Shah, Victoria Krakovna, Daniel Filan, Steven Wang, Dawn Song, Sam Toyer and Dan Hendrycks for their suggestion and helpful feedback on earlier drafts of this paper. We thank Chris Northwood for assistance developing the website accompanying this paper. We are also grateful to our anonymous reviewers for valuable feedback and encouragement to explore defenses in this paper.
\else
Removed for double-blind submission.
\fi
\makeatother

\bibliography{refs}

\clearpage
\appendix
\counterwithin{table}{section}
\counterwithin{figure}{section}
\input{appendix}

\end{document}

%% file: intro.tex
\section{Introduction}

The discovery of adversarial examples for image classifiers prompted a new field of research into adversarial attacks and defenses~\parencite{szegedy:2014}.
Recent work has shown that deep RL policies are also vulnerable to adversarial perturbations of image observations~\parencite{huang:2017,kos:2017}.
However, real-world RL agents inhabit natural environments populated by other agents, including humans, who can only modify another agent's observations via their actions.
We explore whether it's possible to attack a victim policy by building an \textit{adversarial policy} that takes actions in a shared environment, inducing \textit{natural} observations which have adversarial effects on the victim.

RL has been applied in settings as varied as autonomous driving~\parencite{dosovitskiy:2017}, negotiation~\parencite{lewis:2017} and automated trading~\parencite{noonan:2017}.
In domains such as these, an attacker cannot usually directly modify the victim policy's input.
For example, in autonomous driving pedestrians and other drivers can take actions in the world that affect the camera image, but only in a physically realistic fashion.
They cannot add noise to arbitrary pixels, or make a building disappear.
Similarly, in financial trading an attacker can send orders to an exchange which will appear in the victim's market data feed, but the attacker cannot modify observations of a third party's orders.

\begin{figure}
	\centering
	\begin{tabular}{@{}l@{\hskip 0.5em} c}
	\rotatebox[origin=c]{90}{\textbf{Normal}}
	&
	\includegraphics[valign=m,width=0.234\textwidth]{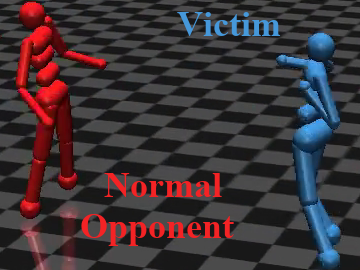}
	\includegraphics[valign=m,width=0.234\textwidth]{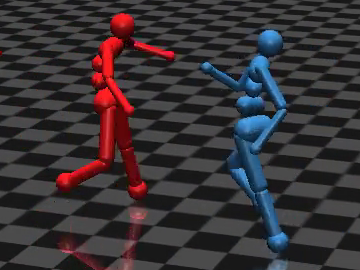}
	\includegraphics[valign=m,width=0.234\textwidth]{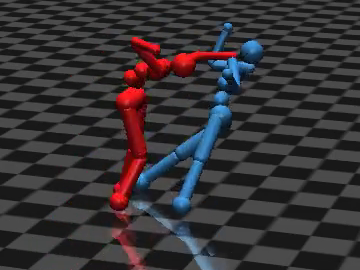}
	\includegraphics[valign=m,width=0.234\textwidth]{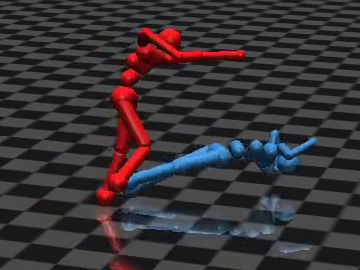}
	\\[3.6em]
	\rotatebox[origin=c]{90}{\textbf{Adversarial}}
	&
	\includegraphics[valign=m,width=0.234\textwidth]{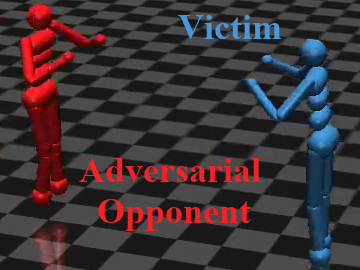}
	\includegraphics[valign=m,width=0.234\textwidth]{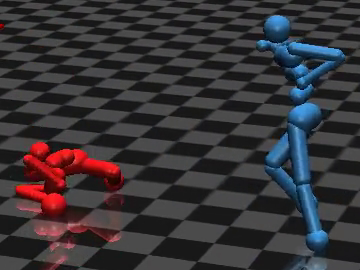}
	\includegraphics[valign=m,width=0.234\textwidth]{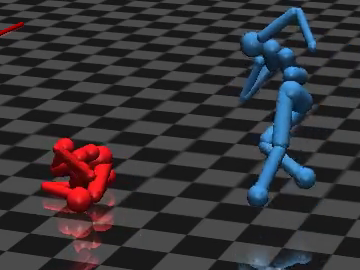}
	\includegraphics[valign=m,width=0.234\textwidth]{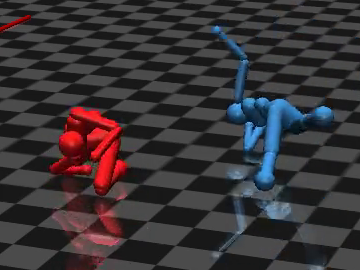}
	\end{tabular}
	\caption{Illustrative snapshots of a victim (in blue) against normal and adversarial opponents (in red). The victim wins if it crosses the finish line; otherwise, the opponent wins. Despite never standing up, the adversarial opponent wins 86\% of episodes, far above the normal opponent's 47\% win rate.}
	\label{fig:intro:snapshots}
\end{figure}

As a proof of concept, we show the existence of adversarial policies in zero-sum simulated robotics games with proprioceptive observations~\parencite{bansal:2018}.
The state-of-the-art victim policies were trained via self-play to be robust to opponents.
We train each adversarial policy using model-free RL against a fixed black-box victim.
We find the adversarial policies reliably beat their victim, despite training for less than 3\% of the timesteps initially used to train the victim policies.

Critically, we find the adversaries win by creating natural observations that are adversarial, and not by becoming generally strong opponents.
Qualitatively, the adversaries fall to the ground in contorted positions, as illustrated in Figure~\ref{fig:intro:snapshots}, rather than learning to run, kick or block like normal opponents.
This strategy does not work when the victim is `masked' and cannot see the adversary's position, suggesting that the adversary succeeds by manipulating a victim's observations through its actions.

Having observed these results, we wanted to understand the sensitivity of the attack to the dimensionality of the victim's observations.
We find that victim policies in higher-dimensional Humanoid environments are substantially more vulnerable to adversarial policies than in lower-dimensional Ant environments.
To gain insight into why adversarial policies succeed, we analyze the activations of the victim's policy network using a Gaussian Mixture Model and t-SNE~\parencite{maaten:2008}.
We find adversarial policies induce significantly different activations than normal opponents, and that the adversarial activations are typically more widely dispersed between timesteps than normal activations.

A natural defense is to fine-tune the victim against the adversary.
We find this protects against that particular adversary, but that repeating the attack method finds a new adversary the fine-tuned victim is vulnerable to.
However, this new adversary differs qualitatively by physically interfering with the victim.
This suggests repeated fine-tuning might provide protection against a range of adversaries.

Our paper makes three contributions. First, we propose a novel, physically realistic threat model for adversarial examples in RL. Second, we demonstrate the existence of adversarial policies in this threat model for several simulated robotics games. Our adversarial policies reliably beat the victim, despite training with less than 3\% as many timesteps and generating seemingly random behavior. Third, we conduct a detailed analysis of why the adversarial policies work. We show they create natural observations that are adversarial to the victim and push the activations of the victim's policy network off-distribution. Additionally, we find policies are easier to attack in high-dimensional environments.

As deep RL is increasingly deployed in environments with potential adversaries, we believe it is important that practitioners are aware of this previously unrecognized threat model.
Moreover, even in benign settings, we believe adversarial policies can be a useful tool for uncovering unexpected policy failure modes.
Finally, we are excited by the potential of adversarial training using adversarial policies, which could improve robustness relative to conventional self-play by training against adversaries that exploit weaknesses undiscovered by the distribution of similar opponents present during self-play.

%% file: related_work.tex
\section{Related Work}

Most study of adversarial examples has focused on small $\ell_p$ norm perturbations to images, which \textcite{szegedy:2014} found cause misclassifications in a variety of models, even though the changes are visually imperceptible to a human.
\textcite{gilmer:2018rules} argue that attackers are not limited to small perturbations, and can instead construct new images or search for naturally misclassified images.
Similarly, \textcite{uesato:2018} argue that the $\ell_p$ model is merely a convenient local approximation for the true worst-case risk.
We follow \textcite{goodfellow:2017} in viewing adversarial examples as any input ``that an attacker has intentionally designed to cause the model to make a mistake.''

Most prior work studying adversarial examples in RL has assumed an $\ell_p$-norm threat model.
\textcite{huang:2017}, \textcite{kos:2017} and \textcite{lin:2017} showed that deep RL policies are vulnerable to small perturbations in image observations.
Our work differs from these previous approaches by using a physically realistic threat model that disallows direct modification of the victim's observations.

A similar threat model was used in \textcite{behzadan:2019}, who pit autonomous vehicles against an adversarial car.
However, in collaborative games like driving, even the optimal policy may be exploitable: we therefore focus on zero-sum games.
\textcite{lanctot:2017} showed agents may be tightly coupled to the agents they were trained with, causing seemingly strong polices to fail against new opponents.
However, the agents we attack beat a range of opponents, and so are not coupled.

Adversarial training is a common defense to adversarial examples, achieving state-of-the-art robustness in image classification~\parencite{xie:2019}.
Prior work has also applied adversarial training to improve the robustness of deep RL policies, where the adversary exerts a force vector on the victim or varies dynamics parameters such as friction~\parencite{pinto:2017,mandlekar:2017,pattanaik:2018}.
Our defense of fine-tuning the victim against the adversary is inspired by this work.

This work follows a rich tradition of worst-case analysis in RL.
In robust MDPs, the transition function is chosen adversarially from an \textit{uncertainty set}~\parencite{bagnell:2001,tamar:2014}.
\textcite{doyle:1996} solve the converse problem: finding the set of transition functions for which a policy is optimal.
Methods also exist to verify controllers or find a counterexample to a specification.
\textcite{bastani:2018} verify decision trees distilled from RL policies, while \textcite{ghosh:2018} test black-box closed-loop simulations.
\defcitealias{ravanbakhsh:2016}{Ravanbakhsh et al (2016)}  %
\citetalias{ravanbakhsh:2016} can even synthesize controllers robust to adversarial disturbances.
Unfortunately, these techniques are only practical in simple environments with low-dimensional adversarial disturbances.
By contrast, while our method lacks formal guarantees, it can test policies in complex multi-agent tasks and naturally scales with improvements in RL algorithms.

%% file: framework.tex
\section{Framework}
\label{sec:framework}

We model the victim as playing against an opponent in a two-player Markov game~\parencite{shapley:1953}.
Our threat model assumes the attacker can control the opponent, in which case we call the opponent an adversary.
We denote the adversary and victim by subscript $\adversary{}$ and $\victim{}$ respectively.
The game $\markovprocess{} = (\statespace, (\actionspace{\adversary}, \actionspace{\victim}), \transition, (\reward{\adversary}, \reward{\victim}))$ consists of state set $\statespace{}$, action sets $\actionspace{\adversary}$ and $\actionspace{\victim}$, and a joint state transition function $\transition{} : \statespace{}\times \actionspaces{} \to \probabilityspace{\statespace{}}$ where $\probabilityspace{\statespace{}}$ is a probability distribution on $\statespace{}$. 
The reward function $\reward{i} : \statespace{} \times \actionspaces{} \times \statespace \to \real$ for player $i \in \{\adversary, \victim\}$ depends on the current state, next state and both player's actions.
Each player wishes to maximize their (discounted) sum of rewards.

The adversary is allowed unlimited black-box access to actions sampled from $\pi_v$, but is not given any white-box information such as weights or activations.
We further assume the victim agent follows a fixed stochastic policy $\pi_v$, corresponding to the common case of a pre-trained model deployed with static weights.
Note that in safety critical systems, where attacks like these would be most concerning, it is standard practice to validate a model and then freeze it, so as to ensure that the deployed model does not develop any new issues due to retraining.
Therefore, a fixed victim is a realistic reflection of what we might see with RL-trained policies in real-world settings, such as with autonomous vehicles.

Since the victim policy $\policy_{\victim}$ is held fixed, the two-player Markov game $\markovprocess{}$ reduces to a single-player MDP $\markovprocess{\adversary} = (\statespace, \actionspace{\adversary}, \transition_{\adversary}, \reward{\adversary}')$ that the attacker must solve.
The state and action space of the adversary are the same as in $\markovprocess{}$, while the transition and reward function have the victim policy $\policy_{\victim}$ embedded:
\begin{equation*}
\transition_{\adversary}\left(\state, \action{\adversary}\right) = \transition\left(\state, \action{\adversary}, \action{\victim}\right)\qquad \text{and} \qquad
\reward{\adversary}'(\state, \action{\adversary}, \nextstate) = \reward{\adversary}(\state, \action{\adversary}, \action{\victim}, \nextstate),
\end{equation*}
where the victim's action is sampled from the stochastic policy $\action{\victim} \sim \policy_{\victim}(\cdot \mid \state)$. The goal of the attacker is to find an adversarial policy $\policy_{\adversary}$ maximizing the sum of discounted rewards:
\begin{equation}
\label{eq:adversary-mdp:objective}
\sum_{t=0}^{\infty} \gamma^t \reward{\adversary}(\state^{(t)},\action{\adversary}^{(t)}, \state^{(t+1)}),\quad \text{where } \state^{(t+1)} \sim \transition_{\adversary}(\state^{(t)}, \action{\adversary}^{(t)})\text{ and }\action{\adversary} \sim \policy_{\adversary}(\cdot \mid \state^{(t)}).
\end{equation}
Note the MDP's dynamics $\transition_{\adversary}$ will be unknown even if the Markov game's dynamics $\transition$ are known since the victim policy $\policy_{\victim}$ is a black-box.
Consequently, the attacker must solve an RL problem.

%% file: bansal_attack.tex
\begin{figure}
	\centering
	\begin{subfigure}{0.24\textwidth}
		\includegraphics[width=\textwidth]{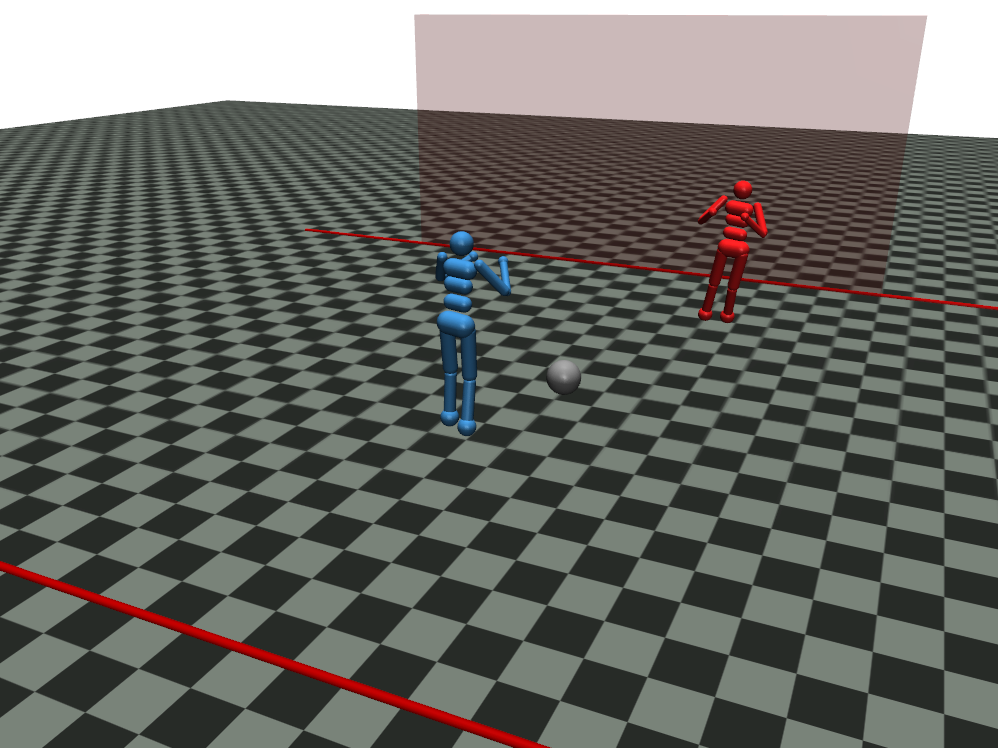}
		\caption{Kick and Defend}
	\end{subfigure}
	\begin{subfigure}{0.24\textwidth}
		\includegraphics[width=\textwidth]{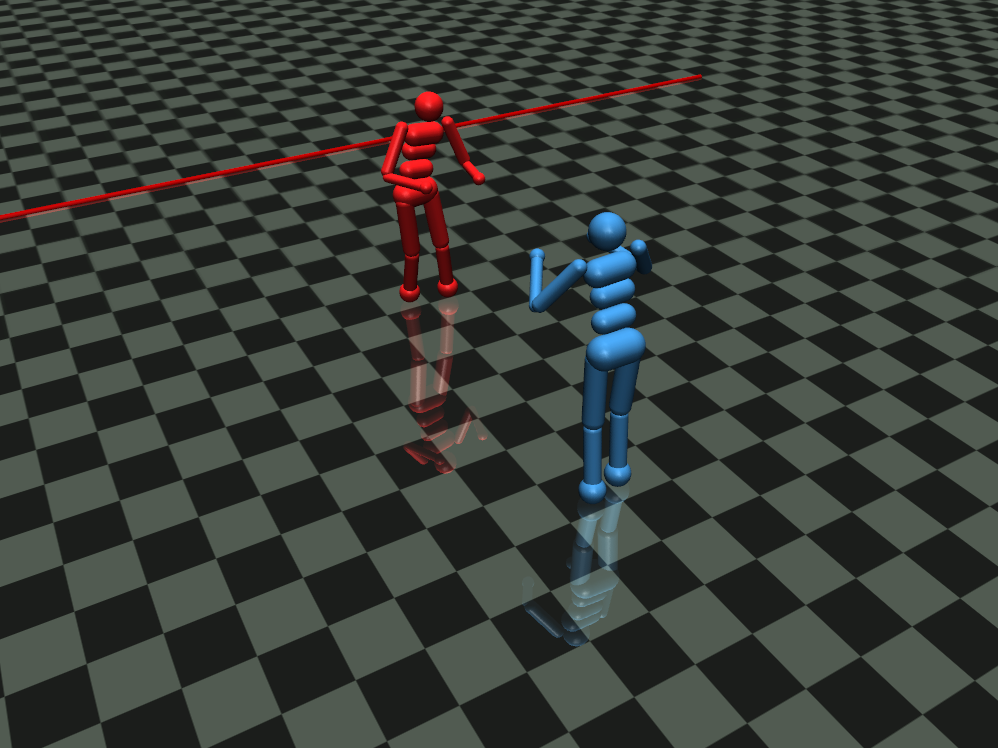}
		\caption{You Shall Not Pass}
	\end{subfigure}
	\begin{subfigure}{0.24\textwidth}
		\includegraphics[width=\textwidth]{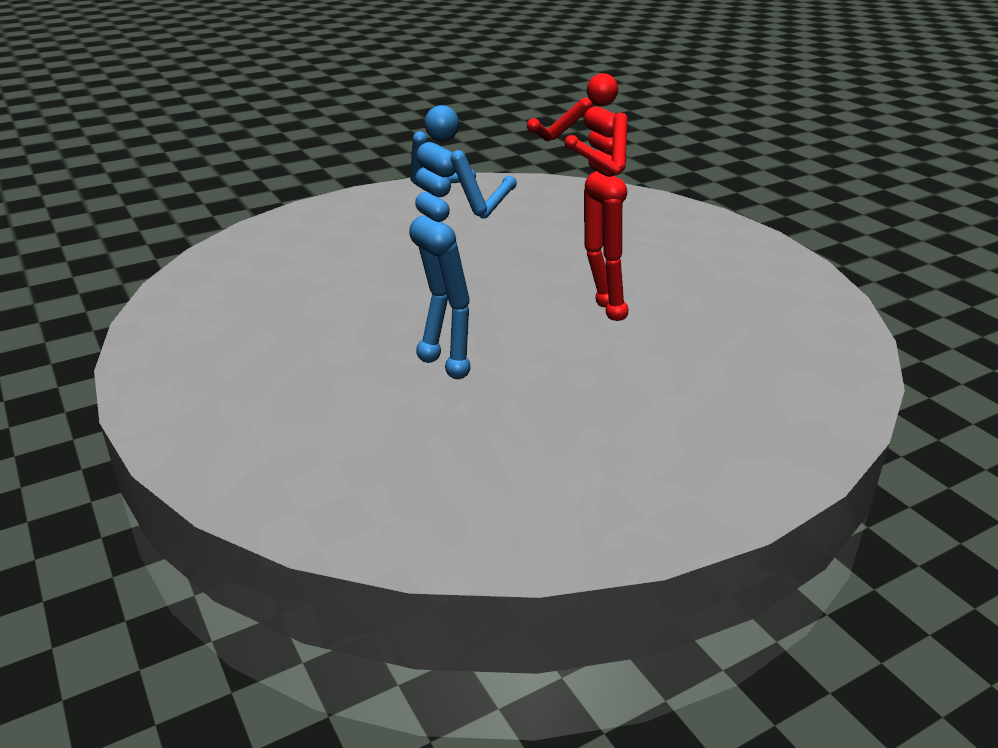}
		\caption{Sumo Humans}
	\end{subfigure}
	\begin{subfigure}{0.24\textwidth}
		\includegraphics[width=\textwidth]{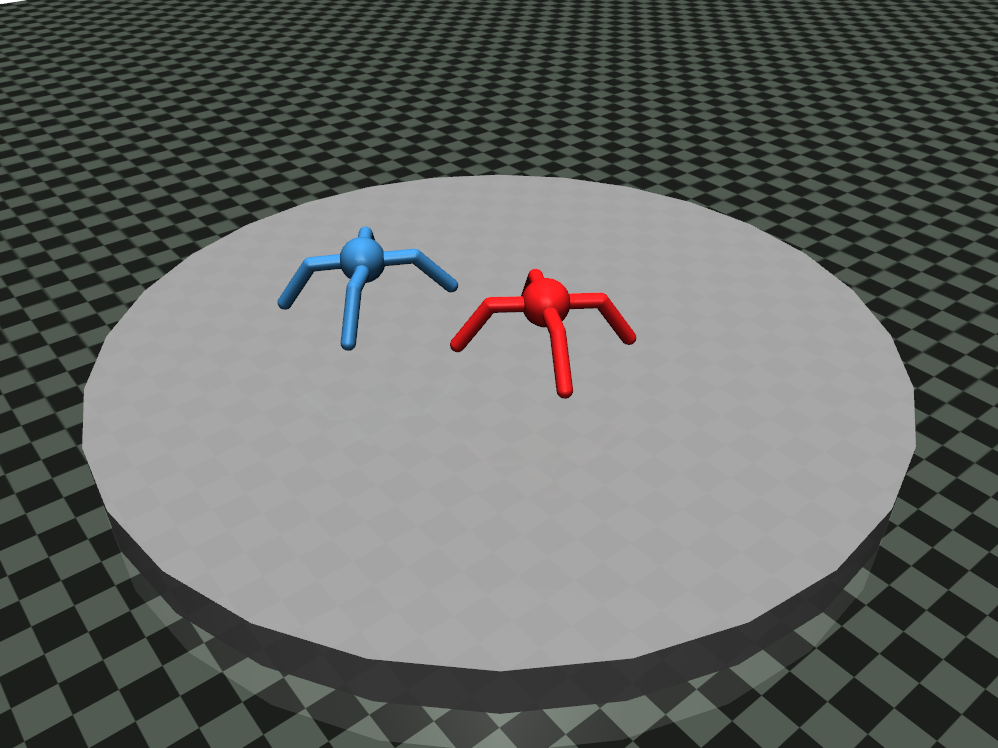}
		\caption{Sumo Ants}
	\end{subfigure}
	\caption{Illustrations of the zero-sum simulated robotics games from \textcite{bansal:2018} we use for evaluation. Environments are further described in Section~\ref{sec:bansal_attack:environments}.}
	\label{fig:environments}
\end{figure}

\section{Finding Adversarial Policies}
We demonstrate the existence of adversarial policies in zero-sum simulated robotics games.
First, we describe how the victim policies were trained and the environments they operate in.
Subsequently, we provide details of our attack method in these environments, and describe several baselines.
Finally, we present a quantitative and qualitative evaluation of the adversarial policies and baseline opponents.

\subsection{Environments and Victim Policies}
\label{sec:bansal_attack:environments}

We attack victim policies for the zero-sum simulated robotics games created by \textcite{bansal:2018}, illustrated in Figure~\ref{fig:environments}.
The victims were trained in pairs via self-play against random old versions of their opponent, for between 680 and 1360 million timesteps.
We use the pre-trained policy weights released in the ``agent zoo'' of \textcite{bansalcode:2018}.
In symmetric environments, the zoo agents are labeled \texttt{ZooN} where $N$ is a random seed.
In asymmetric environments, they are labeled \texttt{ZooVN} and \texttt{ZooON} representing the \textbf{V}ictim and \textbf{O}pponent agents.

All environments are two-player games in the MuJoCo robotics simulator. Both agents observe the position, velocity and contact forces of joints in their body, and the position of their opponent's joints. The episodes end when a win condition is triggered, or after a time limit, in which case the agents draw. We evaluate in all environments from \textcite{bansal:2018} except for \textit{Run to Goal}, which we omit as the setup is identical to \textit{You Shall Not Pass} except for the win condition. We describe the environments below, and specify the number of trained zoo policies and their type (MLP or LSTM):

\textbf{Kick and Defend} (3, LSTM). A soccer penalty shootout between two Humanoid robots. The positions of the kicker, goalie and ball are randomly initialized. The kicker wins if the ball goes between the goalposts; otherwise, the goalie wins, provided it remains within 3 units of the goal.

\textbf{You Shall Not Pass} (1, MLP). Two Humanoid agents are initialized facing each other. The runner wins if it reaches the finish line; the blocker wins if it does not.

\textbf{Sumo Humans} (3, LSTM). Two Humanoid agents compete on a round arena. The players' positions are randomly initialized. A player wins by remaining standing after their opponent has fallen.\footnote{\textcite{bansal:2018} consider the episode to end in a tie if a player falls before it is touched by an opponent. Our win condition allows for attacks that indirectly modify observations without physical contact.}

\textbf{Sumo Ants} (4, LSTM). The same task as \textit{Sumo Humans}, but with `Ant' quadrupedal robot bodies. We use this task in Section~\ref{sec:understanding:dimensionality} to investigate the importance of dimensionality to this attack method.

\subsection{Methods Evaluated}
\label{sec:attack:methods}

Following the RL formulation in Section~\ref{sec:framework}, we train an adversarial policy to maximize Equation~\ref{eq:adversary-mdp:objective} using Proximal Policy Optimization~(PPO; \citealp{schulman:2017}).
We give a sparse reward at the end of the episode, positive when the adversary wins the game and negative when it loses or ties.
\textcite{bansal:2018} trained the victim policies using a similar reward, with an additional dense component at the start of training.
We train for 20 million timesteps using the PPO implementation from Stable Baselines~\parencite{stable-baselines}.
The hyperparameters were selected through a combination of manual tuning and a random search of 100 samples; see Section~\supref{sec:appendix:training} in the appendix for details.
We compare our methods to three baselines: a policy \texttt{Rand} taking random actions; a lifeless policy \texttt{Zero} that exerts zero control; and all pre-trained policies \texttt{Zoo*} from \textcite{bansal:2018}.

\begin{figure}
	\centering
	\includegraphics[width=\textwidth]{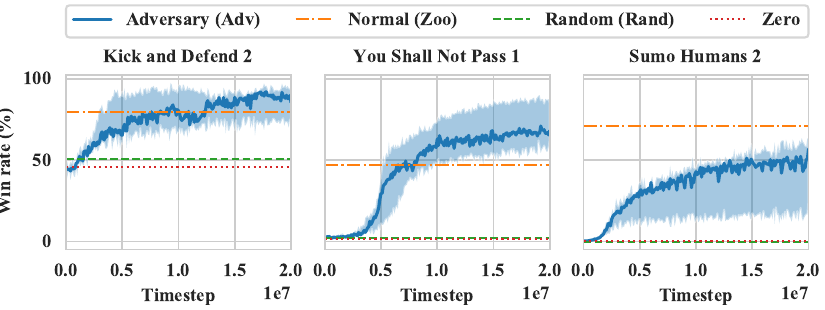}
	\caption{\winratecaption{ against the median victim in each environment (based on the difference between the win rate for \texttt{Adv} and \texttt{Zoo}). The adversary outperforms the \texttt{Zoo} baseline against the median victim in \textit{Kick and Defend} and \textit{You Shall Not Pass}, and is competitive on \textit{Sumo Humans}. For full results, see figure~\ref{fig:score-heatmap} below or figure~\supref{fig:full-win-rate} in the supplementary material.\\ \\}}
	\label{fig:win-rate}
\end{figure}

\subsection{Results}

\begin{figure}
	\centering
	\begin{subfigure}{0.49\textwidth}
		\includegraphics{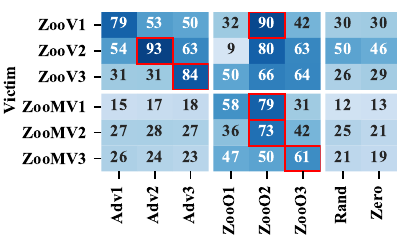}
		\captionsetup{justification=centering}
    \caption{Kick and Defend.}
	\end{subfigure}
	\begin{subfigure}{0.49\textwidth}
		\includegraphics{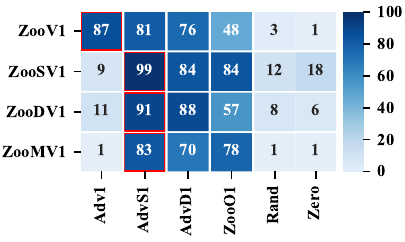}
		\captionsetup{justification=centering}
    \caption{You Shall Not Pass.}
		\label{fig:score-heatmap:ysnp}
	\end{subfigure}
	\begin{subfigure}{0.49\textwidth}
		\includegraphics{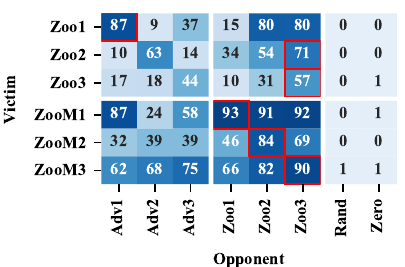}
		\captionsetup{justification=centering}
    \caption{Sumo Humans.}
		\label{fig:score-heatmap:sumohumans}
	\end{subfigure}
	\begin{subfigure}{0.49\textwidth}
		\includegraphics{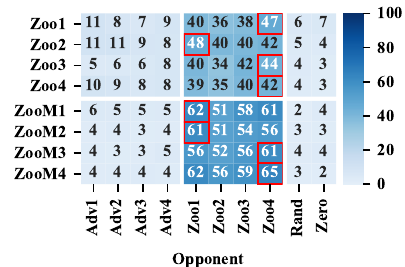}
		\captionsetup{justification=centering}
    \caption{Sumo Ants.}
		\label{fig:score-heatmap:sumoants}
	\end{subfigure}
  \caption{Percentage of games won by opponent (out of 1000), the maximal cell in each row is in red. \scoreagentcaption{}}
	\label{fig:score-heatmap}
\end{figure}

\textbf{Quantitative Evaluation}\ \
We find the adversarial policies reliably win against most victim policies, and outperform the pre-trained \texttt{Zoo} baseline for a majority of environments and victims.
We report the win rate over time against the median victim in each environment in Figure~\ref{fig:win-rate}, with full results in Figure~\supref{fig:full-win-rate} in the supplementary material.
Win rates against all victims are summarized in Figure~\ref{fig:score-heatmap}.

\textbf{Qualitative Evaluation}\ \
The adversarial policies beat the victim not by performing the intended task (e.g. blocking a goal), but rather by exploiting weaknesses in the victim's policy.
This effect is best seen by watching the videos at \website.
In \textit{Kick and Defend} and \textit{You Shall Not Pass}, the adversarial policy never stands up. 
The adversary instead wins by positioning their body to induce adversarial observations that cause the victim's policy to take poor actions.
A robust victim could easily win, a result we demonstrate in Section~\ref{sec:understanding:masking}.

This flavor of attacks is impossible in Sumo Humans, since the adversarial policy immediately loses if it falls over.
Faced with this control constraint, the adversarial policy learns a more high-level strategy: it kneels in the center in a stable position.
Surprisingly, this is very effective against victim 1, which in 88\% of cases falls over attempting to tackle the adversary.
However, it proves less effective against victims 2 and 3, achieving only a 62\% and 45\% win rate, below \texttt{Zoo} baselines.
We further explore the importance of the number of dimensions the adversary can safely manipulate in Section~\ref{sec:understanding:dimensionality}.

\textbf{Distribution Shift}\ \
One might wonder if the adversarial policies win because they are outside the training distribution of the victim.
To test this, we evaluate victims against two simple off-distribution baselines: a random policy \texttt{Rand} (green) and a lifeless policy \texttt{Zero} (red).
These baselines win as often as 30\% to 50\% in \textit{Kick and Defend}, but less than 1\% of the time in \textit{Sumo} and \textit{You Shall Not Pass}.
This is well below the performance of our adversarial policies.
We conclude that most victim policies are robust to off-distribution observations that are not adversarially optimized.

\begin{figure}
	\centering
	\begin{subfigure}[t]{0.48\textwidth}
		\includegraphics[width=\textwidth]{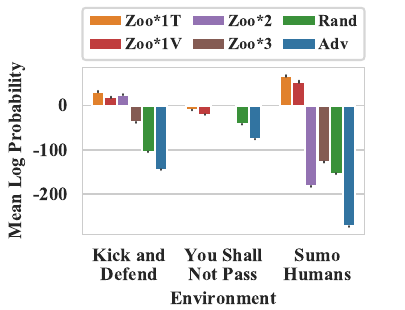}
    \caption{Gaussian Mixture Model (GMM): likelihood the activations of a victim's policy network are ``normal''. We collect activations for $20,000$ timesteps of victim \texttt{Zoo[V]1} playing against each opponent. We fit a $20$-component GMM to activations induced by \texttt{Zoo[O]1}. Error bars are a 95\% confidence interval.}
		\label{fig:understanding:activations:density}
	\end{subfigure}
	\hspace{1em}
 	\begin{subfigure}[t]{0.48\textwidth}
		\includegraphics[width=\textwidth]{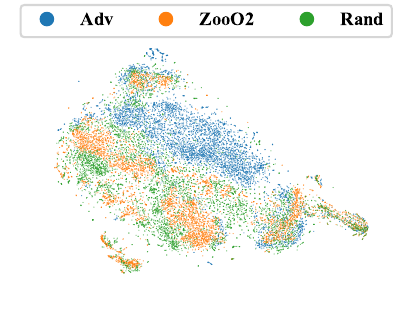}
		\caption{t-SNE activations of Kick and Defend victim \texttt{ZooV2} playing against different opponents. \tsnehyperparams{} See Figures~\supref{fig:tsne:full} and~\supref{fig:tsne:one-opponent} in the supplementary results for visualizations of other environments and victims.}
		\label{fig:understanding:activations:tsne}
	\end{subfigure}
	\caption{Analysis of activations of the victim's policy network. Both figures show the adversary \texttt{Adv} induces off-distribution activations. \textbf{Key}: legends specify opponent the victim played against. \texttt{Adv} is the best adversary trained against the victim, and \texttt{Rand} is a policy taking random actions. \texttt{Zoo*N} corresponds to \texttt{ZooN} (Sumo) or \texttt{ZooON} (otherwise). \texttt{Zoo*1T} and \texttt{Zoo*1V} are the train and validation datasets, drawn from \texttt{Zoo1} (Sumo) or \texttt{ZooO1} (otherwise).}
\label{fig:understanding:activations}
\end{figure}

%% file: understanding.tex
\section{Understanding Adversarial Policies}

In the previous section we demonstrated adversarial policies exist for victims in a range of competitive simulated robotics environments. 
In this section, we focus on understanding why these policies exist.
Specifically, we establish that adversarial policies manipulate the victim through their body position; that victims are more vulnerable to adversarial policies in high-dimensional environments; and that activations of the victim's policy network differ substantially when playing an adversarial opponent.

\subsection{Masked Policies}
\label{sec:understanding:masking}

We have previously shown that adversarial policies are able to reliably win against victims.
In this section, we demonstrate that they win by taking actions to induce natural observations that are adversarial to the victim, and not by physically interfering with the victim.
To test this, we introduce a `masked` victim (labeled \texttt{ZooMN} or \texttt{ZooMVN}) that is the same as the normal victim \texttt{ZooN} or \texttt{ZooVN}, except the observation of the adversary's position is set to a static value corresponding to a typical initial position.
We use the same adversarial policy against the normal and masked victim.

One would expect it to be beneficial to be able to see your opponent.
Indeed, the masked victims do worse than a normal victim when playing normal opponents.
For example, Figure~\ref{fig:score-heatmap:ysnp} shows that in \textit{You Shall Not Pass} the normal opponent \texttt{ZooO1} wins 78\% of the time against the masked victim \texttt{ZooMV1} but only 47\% of the time against the normal victim \texttt{ZooV1}.
However, the relationship is reversed when playing an adversary.
The normal victim \texttt{ZooV1} loses 86\% of the time to adversary \texttt{Adv1} whereas the masked victim \texttt{ZooMV1} wins 99\% of the time.
This pattern is particularly clear in \textit{You Shall Not Pass}, but the trend is similar in other environments.

This result is surprising as it implies highly non-transitive relationships may exist between policies even in games that seem to be transitive.
A game is said to be transitive if policies can be ranked such that higher-ranked policies beat lower-ranked policies.
Prima facie, the games in this paper seem transitive: professional human soccer players and sumo wrestlers can reliably beat amateurs.
Despite this, there is a non-transitive relationship between adversarial policies, victims and masked victims. 
Consequently, we urge caution when using methods such as self-play that assume transitivity, and would recommend more general methods where practical~\parencite{balduzzi:2019,brown:2019}.

Our findings also suggest a trade-off in the size of the observation space.
In benign environments, allowing more observation of the environment increases performance.
However, this also makes the agent more vulnerable to adversaries.
This is in contrast to an idealized Bayesian agent, where the value of information is always non-negative~\parencite{good:1967}.
In the following section, we investigate further the connection between vulnerability to attack and the size of the observation space.

\subsection{Dimensionality}
\label{sec:understanding:dimensionality}

A variety of work has concluded that classifiers are more vulnerable to adversarial examples on high-dimensional inputs~\parencite{gilmer:2018spheres,khoury:2018,shafahi:2019}.
We hypothesize a similar result for RL policies: the greater the dimensionality of the component $P$ of the observation space under control of the adversary, the more vulnerable the victim is to attack.
We test this hypothesis in the Sumo environment, varying whether the agents are Ants or Humanoids.
The results in Figures~\ref{fig:score-heatmap:sumohumans} and~\ref{fig:score-heatmap:sumoants} support the hypothesis.
The adversary has a much lower win-rate in the low-dimensional \textit{Sumo Ants} ($\dim P = 15$) environment than in the higher dimensional \textit{Sumo Humans} ($\dim P = 24$) environment, where $P$ is the position of the adversary's joints.

\subsection{Victim Activations}
\label{sec:understanding:tsne}

In Section~\ref{sec:understanding:masking} we showed that adversarial policies win by creating natural observations that are adversarial to the victim.
In this section, we seek to better understand \textit{why} these observations are adversarial.
We record activations from each victim's policy network playing a range of opponents, and analyze these using a Gaussian Mixture Model (GMM) and a t-SNE visualization.
See Section~\ref{sec:appendix:activations} in the supplementary material for details of training and hyperparameters.

We fit a GMM on activations \texttt{Zoo*1T} collected playing against a normal opponent, \texttt{Zoo1} or \texttt{ZooV1}, holding out \texttt{Zoo*1V} for validation.
Figure~\ref{fig:understanding:activations:density} shows that the adversarial policy \texttt{Adv} induces activations with the lowest log-likelihood, with random baseline \texttt{Rand} only slightly more probable.
Normal opponents \texttt{Zoo*2} and \texttt{Zoo*3} induce activations with almost as high likelihood as the validation set \texttt{Zoo*1V}, except in \textit{Sumo Humans} where they are as unlikely as \texttt{Rand}.

We plot a t-SNE visualization of the activations of Kick and Defend victim \texttt{ZooV2} in Figure~\ref{fig:understanding:activations:tsne}.
As expected from the density model results, there is a clear separation between between \texttt{Adv}, \texttt{Rand} and the normal opponent \texttt{ZooO2}.
Intriguingly, \texttt{Adv} induces activations more widely dispersed than the random policy \texttt{Rand}, which in turn are more widely dispersed than \texttt{ZooO2}.
We report on the full set of victim policies in Figures~\supref{fig:tsne:full} and \supref{fig:tsne:one-opponent} in the supplementary material.

%% file: defenses.tex
\section{Defending Against Adversarial Policies}

The ease with which policies can be attacked highlights the need for effective defenses.
A natural defense is to fine-tune the victim zoo policy against an adversary, which we term \textit{single} training.
We also investigate \textit{dual} training, randomly picking either an adversary or a zoo policy at the start of each episode.
The training procedure is otherwise the same as for adversaries, described in Section~\ref{sec:attack:methods}.

We report on the win rates in \textit{You Shall Not Pass} in Figure~\ref{fig:score-heatmap:ysnp}.
We find both the single \texttt{ZooSV1} and dual \texttt{ZooDV1} fine-tuned victims are robust to adversary \texttt{Adv1}, with the adversary win rate dropping from 87\% to around 10\%.
However, \texttt{ZooSV1} catastrophically forgets how to play against the normal opponent \texttt{ZooO1}.
The dual fine-tuned victim \texttt{ZooDV1} fares better, with opponent \texttt{ZooO1} winning only 57\% of the time.
However, this is still an increase from \texttt{ZooO1}'s 48\% win rate against the original victim \texttt{ZooV1}.
This suggests \texttt{ZooV1} may use features that are helpful against a normal opponent but which are easily manipulable~\parencite{ilyas:2019}.

Although the fine-tuned victims are robust to the original adversarial policy \texttt{Adv1}, they are still vulnerable to our attack method.
New adversaries \texttt{AdvS1} and \texttt{AdvD1} trained against \texttt{ZooSV1} and \texttt{ZooDV1} win at equal or greater rates than before, and transfer successfully to the original victim.
However, the new adversaries \texttt{AdvS1} and \texttt{AdvD1} are qualitatively different, tripping the victim up by lying prone on the ground, whereas \texttt{Adv1} causes \texttt{ZooV1} to fall without ever touching it.

%% file: discussion.tex
\section{Discussion}

\textbf{Contributions}.
Our paper makes three key contributions.
\textbf{First}, we have proposed a novel threat model of \textit{natural} adversarial observations produced by an adversarial policy taking actions in a shared environment.
\textbf{Second}, we demonstrate that adversarial policies exist in a range of zero-sum simulated robotics games against state-of-the-art victims trained via self-play to be robust to adversaries.
\textbf{Third}, we verify the adversarial policies win by confusing the victim, not by learning a generally strong policy.
Specifically, we find the adversary induces highly off-distribution activations in the victim, and that victim performance \textit{increases} when it is blind to the adversary's position.

\textbf{Self-play}.
While it may at first appear unsurprising that a policy trained as an adversary against another RL policy would be able to exploit it, we believe that this observation is highly significant.
The policies we have attacked were explicitly trained via self-play to be robust.
Although it is known that self-play with deep RL may not converge, or converge only to a local rather than global Nash, self-play has been used with great success in a number of works focused on playing adversarial games directly against humans~\parencite{silver:2018,openai:2018}.
Our work shows that even apparently strong self-play policies can harbor serious but hard to find failure modes, demonstrating these theoretical limitations are practically relevant and highlighting the need for careful testing.

Our attack provides some amount of testing by constructively lower-bounding the exploitability of a victim policy -- its performance against its worst-case opponent -- by training an adversary.
Since the victim's win rate declines against our adversarial policy, we can confirm that the victim and its self-play opponent were not in a global Nash.
Notably we expect our attack to succeed even for policies in a local Nash, as the adversary is trained starting from a random point that is likely outside the victim's attractive basin.

\textbf{Defense}.
We implemented a simple defense: fine-tuning the victim against the adversary.
We find our attack can be successfully reapplied to beat this defense, suggesting adversarial policies are difficult to eliminate.
However, the defense does appear to protect against attacks that rely on confusing the victim: the new adversarial policy is forced to instead trip the victim up.
We therefore believe that scaling up this defense is a promising direction for future work.
In particular, we envisage a variant of population-based training where new agents are continually added to the pool to promote diversity, and agents train against a fixed opponent for a prolonged period of time to avoid local equilibria.

\textbf{Conclusion}.
Overall, we are excited about the implications the adversarial policy model has for the robustness, security and understanding of deep RL policies.
Our results show the existence of a previously unrecognized problem in deep RL, and we hope this work encourages other researchers to investigate this area further.
Videos and other supplementary material are available online at \website{} and our source code is available on GitHub at \github{}.

%% file: appendix.tex
\section{Training: hyperparameters and computational infrastructure}
\label{sec:appendix:training}

\begin{table}[h]
	\centering
	\begin{tabular}{l l l l}
	\toprule
	Parameter & Value & Search Range & Search Distribution \\
	\midrule
	Total Timesteps & $\num{20e6}$ & $[\num{0}, \num{40e6}]$ & Manual \\
	Batch size & $\num{16384}$ & $[\num{2048}, \num{65536}]$ & Log uniform \\
	Number of environments & 8 & $[\num{1},\num{16}]$ & Manual \\
	Mini-batches & $\num{4}$ & $[\num{1}, \num{128}]$ & Log uniform \\
	Epochs per update & $\num{4}$ & $[\num{1}, \num{11}]$ & Uniform \\
	Learning rate & $\num{3e-4}$ & $[\num{1e-5}, \num{1e-2}]$ & Log uniform \\
	Discount & $0.99$ & --- & --- \\
	Maximum Gradient Norm & $\num{0.5}$ & --- & --- \\
	Clip Range & $\num{0.2}$ & --- & --- \\
	Advantage Estimation Discount & $\num{0.95}$ & --- & --- \\
	Entropy coefficient & $\num{0.0}$ & --- & --- \\
	Value Function Loss Coefficient & $\num{0.5}$ & --- & --- \\
	\bottomrule
	\end{tabular}
	\caption{Hyperparameters for Proximal Policy Optimization.}
	\label{tab:appendix:training:hyperparams}
\end{table}

Table~\ref{tab:appendix:training:hyperparams} specifies the hyperparameters used for training. The number of environments was chosen for performance reasons after observing diminishing returns from using more than 8 parallel environments. The total timesteps was chosen by inspection after observing diminishing returns to additional training. The batch size, mini-batches, epochs per update, entropy coefficient and learning rate were tuned via a random search with 100 samples on two environments, \textit{Kick and Defend} and \textit{Sumo Humans}.  All other hyperparameters are the defaults in the \texttt{PPO2} implementation in Stable Baselines~\parencite{stable-baselines}.

We repeated the hyperparameter sweep for fine-tuning victim policies for the defense experiments, but obtained similar results. For simplicity, we therefore chose to use the same hyperparameters throughout.

We used a mixture of in-house and cloud infrastructure to perform these experiments. It takes around 8 hours to train an adversary for a single victim using 4 cores of an Intel Xeon Platinum 8000 (Skylake) processor.

\section{Activation Analysis: t-SNE and GMM}
\label{sec:appendix:activations}
We collect activations from all feed forward layers of the victim's policy network.
This gives two $64$-length vectors, which we concatenate into a single $128$-dimension vector for analysis with a Gaussian Mixture Model and a t-SNE representation.

\subsection{t-SNE hyperparameter selection}
We fit models with perplexity $5$, $10$, $20$, $50$, $75$, $100$, $250$ and $1000$. We chose $250$ since qualitatively it produced the clearest visualization of data with a moderate number of distinct clusters.

\subsection{Gaussian Mixture Model hyperparameter selection}
We fit models with $5$, $10$, $20$, $40$ and $80$ components with a full (unrestricted) and diagonal covariance matrix. We used the Bayesian Information Criterion (BIC) and average log-likelihood on a held-out validation set as criteria for selecting hyperparameters. We found $20$ components with a full covariance matrix achieved the lowest BIC and highest validation log-likelihood in the majority of environment-victim pairs, and was the runner-up in the remainder.

\section{Figures}
Supplementary figures are provided on the subsequent pages.

\begin{figure}
	\includegraphics{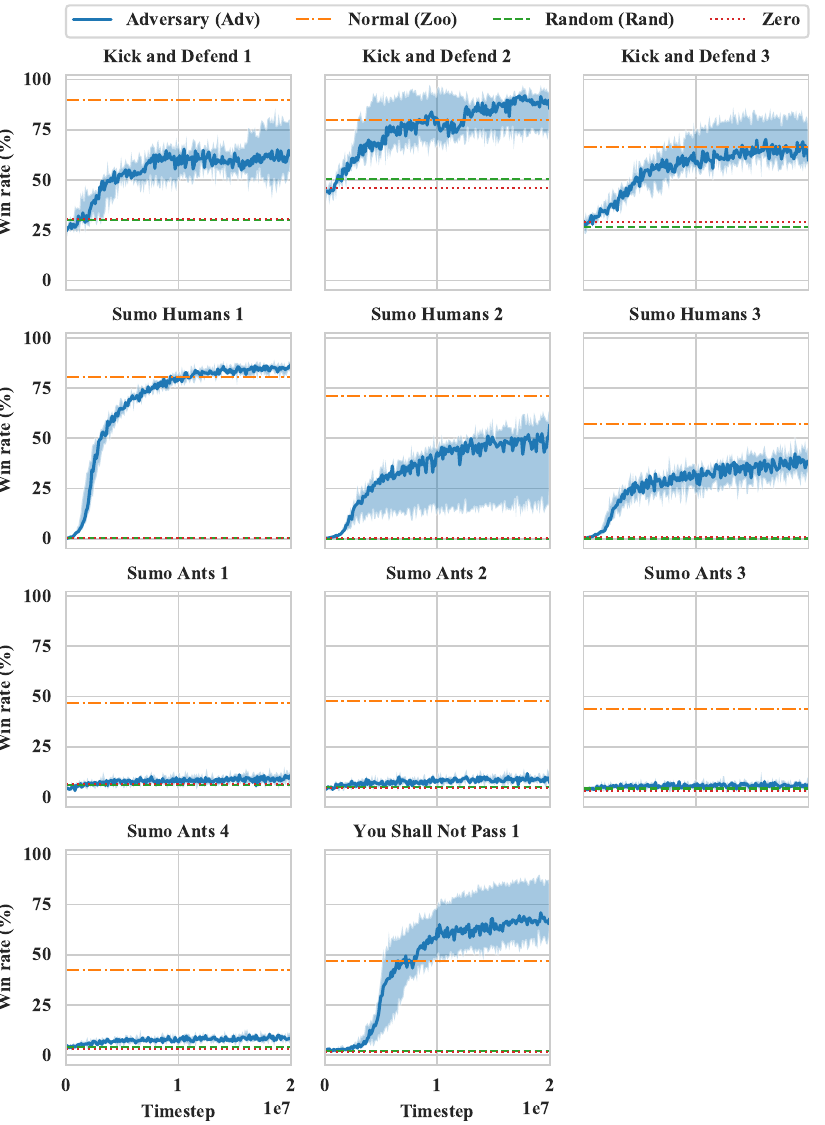}
        \caption{\winratecaption{. The adversary exceeds baseline win rates against most victims in \textit{Kick and Defend} and \textit{You Shall Not Pass}, is competitive on \textit{Sumo Humans}, but performs poorly in the low-dimensional \textit{Sumo Ants} environment. }}
	\label{fig:full-win-rate}
\end{figure}

\begin{figure}
	\begin{subfigure}{\textwidth}
		\includegraphics[width=\textwidth]{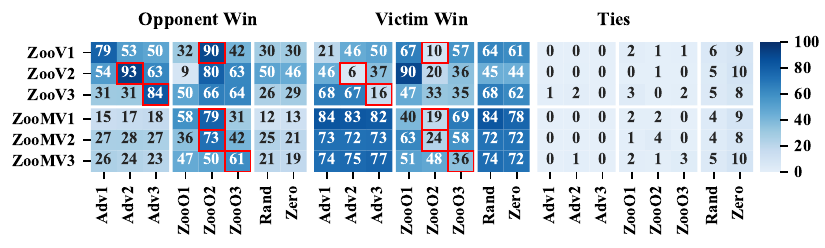}
		\caption{Kick and Defend}
	\end{subfigure}
	\begin{subfigure}{\textwidth}
		\includegraphics[width=\textwidth]{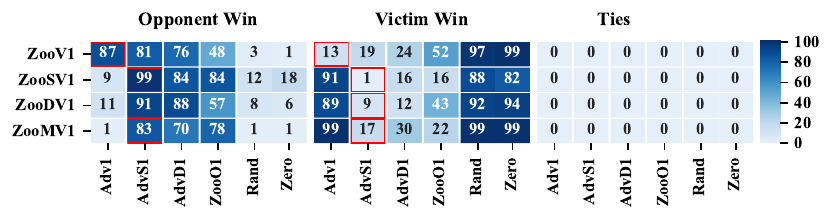}
		\caption{You Shall Not Pass}
	\end{subfigure}
	\begin{subfigure}{\textwidth}
		\includegraphics[width=\textwidth]{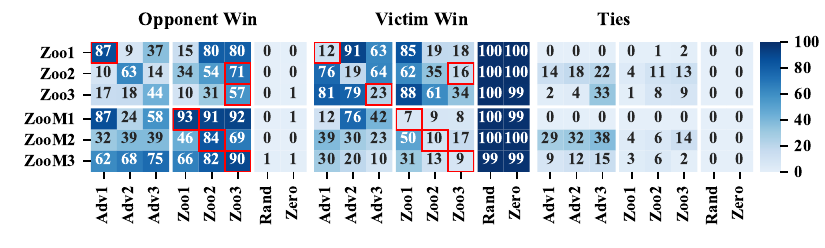}
		\caption{Sumo Humans}
	\end{subfigure}
	\begin{subfigure}{\textwidth}
		\includegraphics[width=\textwidth]{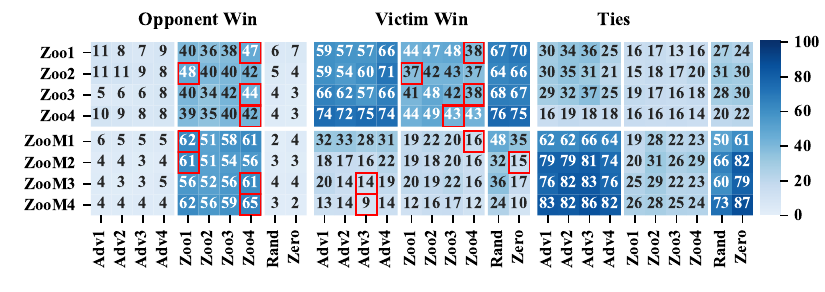}
		\caption{Sumo Ants}
	\end{subfigure}
	\caption{Percentage of episodes (out of 1000) won by the opponent, the victim or tied.
	The maximal opponent win rate in each row is in red.
    Victims are on the $y$-axis and opponents on the $x$-axis.
    \scoreagentcaption{}}
	\label{fig:full-score-heatmap}
\end{figure}

\begin{figure}
	\includegraphics[width=\textwidth]{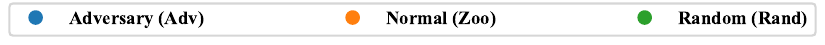}
	\begin{subfigure}{0.33\textwidth}
		\includegraphics[width=\textwidth]{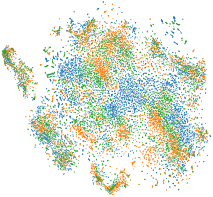}
		\caption{Kick and Defend, victim 1}
	\end{subfigure}
	\begin{subfigure}{0.33\textwidth}
		\includegraphics[width=\textwidth]{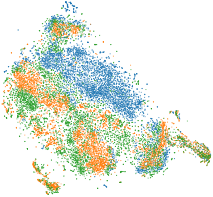}
		\caption{Kick and Defend, victim 2}
	\end{subfigure}
	\begin{subfigure}{0.33\textwidth}
		\includegraphics[width=\textwidth]{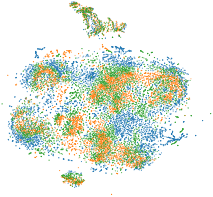}
		\caption{Kick and Defend, victim 3}
	\end{subfigure}
	\begin{subfigure}{0.33\textwidth}
		\includegraphics[width=\textwidth]{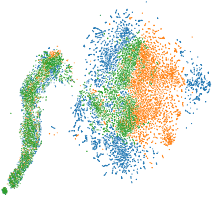}
		\caption{Sumo Humans, victim 1}
	\end{subfigure}
	\begin{subfigure}{0.33\textwidth}
		\includegraphics[width=\textwidth]{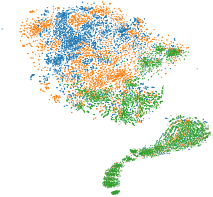}
		\caption{Sumo Humans, victim 2}
	\end{subfigure}
	\begin{subfigure}{0.33\textwidth}
		\includegraphics[width=\textwidth]{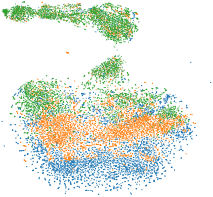}
		\caption{Sumo Humans, victim 3}
	\end{subfigure}
	\begin{subfigure}{0.33\textwidth}
		\includegraphics[width=\textwidth]{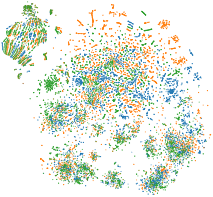}
		\caption{Sumo Ants, victim 1}
	\end{subfigure}
	\begin{subfigure}{0.33\textwidth}
		\includegraphics[width=\textwidth]{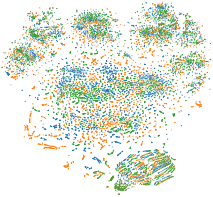}
		\caption{Sumo Ants, victim 2}
	\end{subfigure}
	\begin{subfigure}{0.33\textwidth}
		\includegraphics[width=\textwidth]{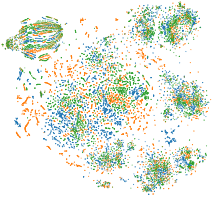}
		\caption{Sumo Ants, victim 3}
	\end{subfigure}
	\begin{subfigure}{0.33\textwidth}
		\includegraphics[width=\textwidth]{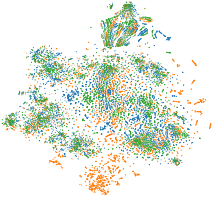}
		\caption{Sumo Ants, victim 4}
	\end{subfigure}
	\begin{subfigure}{0.33\textwidth}
		\includegraphics[width=\textwidth]{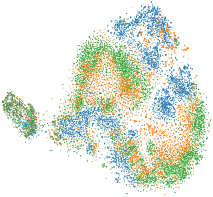}
		\caption{You Shall Not Pass, victim 1}
	\end{subfigure}
	\caption{t-SNE activations of the victim when playing against different opponents. There is a clear separation between the activations induced by \texttt{Adv} and those of the normal opponent \texttt{Zoo}. \tsnehyperparams{} The victim is \texttt{ZooN} (\textit{Sumo}) or \texttt{ZooVN} (other environments), where $N$ is given in the figure caption. Opponent \texttt{Adv} is the best adversary trained against the victim. Opponent \texttt{Zoo} corresponds to \texttt{ZooN} (\textit{Sumo}) or \texttt{ZooON} (other environments). See Figure~\ref{fig:tsne:one-opponent} for activations for a single opponent at a time.}
	\label{fig:tsne:full}
\end{figure}

\begin{figure}
	\includegraphics[width=\textwidth]{figs/tsne/KickAndDefend-v0_victim_zoo_1/external_legend.pdf}
	\begin{subfigure}{\textwidth}
		\includegraphics[width=\textwidth]{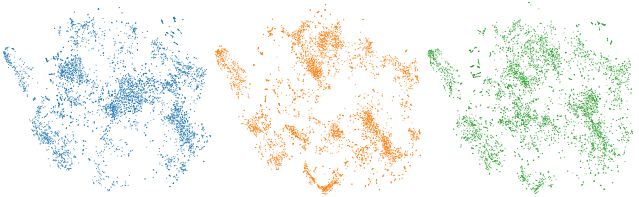}
		\caption{Kick and Defend.}
	\end{subfigure}
	\begin{subfigure}{\textwidth}
		\includegraphics[width=\textwidth]{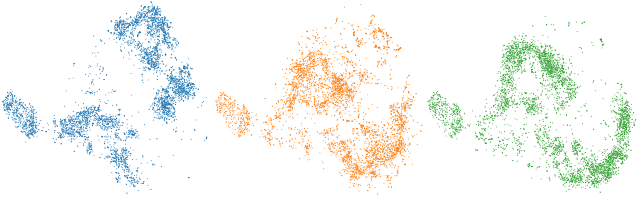}
		\caption{You Shall Not Pass.}
	\end{subfigure}
	\begin{subfigure}{\textwidth}
		\includegraphics[width=\textwidth]{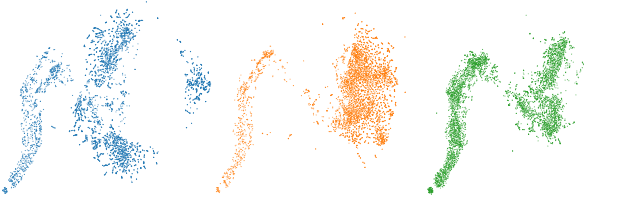}
		\caption{Sumo Humans.}
	\end{subfigure}
	\begin{subfigure}{\textwidth}
		\includegraphics[width=\textwidth]{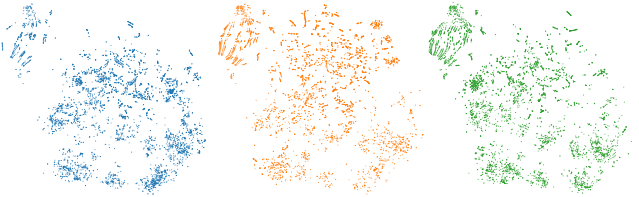}
		\caption{Sumo Ants.}
	\end{subfigure}
	\caption{t-SNE activations of victim \texttt{Zoo1} (\textit{Sumo}) or \texttt{ZooV1} (other environments). The results are the same as in Figure~\ref{fig:tsne:full} but decomposed into individual opponents for clarity. \tsnehyperparams{} Opponent \texttt{Adv} is the best adversary trained against the victim. Opponent \texttt{Zoo} is \texttt{Zoo1} (\textit{Sumo}) or \texttt{ZooO1} (other environments). See Figure~\ref{fig:tsne:full} for results for other victims (one plot per victim).}
	\label{fig:tsne:one-opponent}
\end{figure}

%% file: main.bbl
\begin{thebibliography}{37}
\providecommand{\natexlab}[1]{#1}
\providecommand{\url}[1]{\texttt{#1}}
\expandafter\ifx\csname urlstyle\endcsname\relax
  \providecommand{\doi}[1]{doi: #1}\else
  \providecommand{\doi}{doi: \begingroup \urlstyle{rm}\Url}\fi

\bibitem[Bagnell et~al.(2001)Bagnell, Ng, and Schneider]{bagnell:2001}
J.~Andrew Bagnell, Andrew~Y. Ng, and Jeff~G. Schneider.
\newblock Solving uncertain {Markov} decision processes.
\newblock Technical Report CMU-RI-TR-01-25, August 2001.

\bibitem[Balduzzi et~al.(2019)Balduzzi, Garnelo, Bachrach, Czarnecki,
  P{\'{e}}rolat, Jaderberg, and Graepel]{balduzzi:2019}
David Balduzzi, Marta Garnelo, Yoram Bachrach, Wojciech~M. Czarnecki, Julien
  P{\'{e}}rolat, Max Jaderberg, and Thore Graepel.
\newblock Open-ended learning in symmetric zero-sum games.
\newblock arXiv:1901.08106v1 [cs.LG], 2019.

\bibitem[Bansal et~al.(2018{\natexlab{a}})Bansal, Pachocki, Sidor, Sutskever,
  and Mordatch]{bansal:2018}
Trapit Bansal, Jakub Pachocki, Szymon Sidor, Ilya Sutskever, and Igor Mordatch.
\newblock Emergent complexity via multi-agent competition.
\newblock In \emph{Proceedings of the International Conference on Learning
  Representations (ICLR)}, 2018{\natexlab{a}}.

\bibitem[Bansal et~al.(2018{\natexlab{b}})Bansal, Pachocki, Sidor, Sutskever,
  and Mordatch]{bansalcode:2018}
Trapit Bansal, Jakub Pachocki, Szymon Sidor, Ilya Sutskever, and Igor Mordatch.
\newblock Source code and model weights for emergent complexity via multi-agent
  competition, 2018{\natexlab{b}}.
\newblock URL \url{https://github.com/openai/multiagent-competition}.

\bibitem[Bastani et~al.(2018)Bastani, Pu, and Solar-Lezama]{bastani:2018}
Osbert Bastani, Yewen Pu, and Armando Solar-Lezama.
\newblock Verifiable reinforcement learning via policy extraction.
\newblock In \emph{Advances in Neural Information Processing Systems
  (NeurIPS)}, pages 2494--2504. 2018.

\bibitem[{Behzadan} and {Munir}(2019)]{behzadan:2019}
Vahid {Behzadan} and Arslan {Munir}.
\newblock Adversarial reinforcement learning framework for benchmarking
  collision avoidance mechanisms in autonomous vehicles.
\newblock \emph{IEEE Intelligent Transportation Systems Magazine}, 2019.

\bibitem[Brown et~al.(2019)Brown, Lerer, Gross, and Sandholm]{brown:2019}
Noam Brown, Adam Lerer, Sam Gross, and Tuomas Sandholm.
\newblock Deep counterfactual regret minimization.
\newblock In \emph{Proceedings of the International Conference on Machine
  Learning (ICML)}, 2019.

\bibitem[Dosovitskiy et~al.(2017)Dosovitskiy, Ros, Codevilla, Lopez, and
  Koltun]{dosovitskiy:2017}
Alexey Dosovitskiy, German Ros, Felipe Codevilla, Antonio Lopez, and Vladlen
  Koltun.
\newblock {CARLA}: {An} open urban driving simulator.
\newblock In \emph{Proceedings of the Conference on Robot Learning (CoRL)},
  volume~78, pages 1--16, 2017.

\bibitem[Doyle et~al.(1996)Doyle, Primbs, Shapiro, and
  Nevisti\'{c}]{doyle:1996}
John Doyle, James~A. Primbs, Benjamin Shapiro, and Vesna Nevisti\'{c}.
\newblock Nonlinear games: examples and counterexamples.
\newblock In \emph{Proceedings of IEEE Conference on Decision and Control
  (CDC)}, volume~4, pages 3915--3920, 1996.

\bibitem[Ghosh et~al.(2018)Ghosh, Berkenkamp, Ranade, Qadeer, and
  Kapoor]{ghosh:2018}
Shromona Ghosh, Felix Berkenkamp, Gireeja Ranade, Shaz Qadeer, and Ashish
  Kapoor.
\newblock Verifying controllers against adversarial examples with {Bayesian}
  optimization.
\newblock In \emph{{IEEE} International Conference on Robotics and Automation
  (ICRA)}, pages 7306--7313, 2018.

\bibitem[Gilmer et~al.(2018{\natexlab{a}})Gilmer, Adams, Goodfellow, Andersen,
  and Dahl]{gilmer:2018rules}
Justin Gilmer, Ryan~P. Adams, Ian Goodfellow, David Andersen, and George~E.
  Dahl.
\newblock Motivating the rules of the game for adversarial example research.
\newblock arXiv:1807.06732v2 [cs.LG], 2018{\natexlab{a}}.

\bibitem[Gilmer et~al.(2018{\natexlab{b}})Gilmer, Metz, Faghri, Schoenholz,
  Raghu, Wattenberg, and Goodfellow]{gilmer:2018spheres}
Justin Gilmer, Luke Metz, Fartash Faghri, Samuel~S. Schoenholz, Maithra Raghu,
  Martin Wattenberg, and Ian Goodfellow.
\newblock Adversarial spheres.
\newblock arXiv:1801.02774v3 [cs.CV], 2018{\natexlab{b}}.

\bibitem[Good(1967)]{good:1967}
I.J. Good.
\newblock On the principle of total evidence.
\newblock \emph{The British Journal for the Philosophy of Science}, 17\penalty0
  (4):\penalty0 319--321, 1967.

\bibitem[Goodfellow et~al.(2017)Goodfellow, Papernot, Huang, Duan, Abbeel, and
  Clark]{goodfellow:2017}
Ian Goodfellow, Nicolas Papernot, Sandy Huang, Yan Duan, Pieter Abbeel, and
  Jack Clark.
\newblock Attacking machine learning with adversarial examples.
\newblock \url{https://openai.com/blog/adversarial-example-research/}, 2017.

\bibitem[Hill et~al.(2019)Hill, Raffin, Ernestus, Gleave, Kanervisto, Traore,
  Dhariwal, Hesse, Klimov, Nichol, Plappert, Radford, Schulman, Sidor, and
  Wu]{stable-baselines}
Ashley Hill, Antonin Raffin, Maximilian Ernestus, Adam Gleave, Anssi
  Kanervisto, Rene Traore, Prafulla Dhariwal, Christopher Hesse, Oleg Klimov,
  Alex Nichol, Matthias Plappert, Alec Radford, John Schulman, Szymon Sidor,
  and Yuhuai Wu.
\newblock {Stable Baselines}.
\newblock \url{https://github.com/hill-a/stable-baselines}, 2019.

\bibitem[Huang et~al.(2017)Huang, Papernot, Goodfellow, Duan, and
  Abbeel]{huang:2017}
Sandy~H. Huang, Nicolas Papernot, Ian~J. Goodfellow, Yan Duan, and Pieter
  Abbeel.
\newblock Adversarial attacks on neural network policies.
\newblock arXiv:1702.02284v1 [cs.LG], 2017.

\bibitem[Ilyas et~al.(2019)Ilyas, Santurkar, Tsipras, Engstrom, Tran, and
  Madry]{ilyas:2019}
Andrew Ilyas, Shibani Santurkar, Dimitris Tsipras, Logan Engstrom, Brandon
  Tran, and Aleksander Madry.
\newblock Adversarial examples are not bugs, they are features.
\newblock arXiv:1905.02175v4 [stat.ML], 2019.

\bibitem[Khoury and Hadfield-Menell(2018)]{khoury:2018}
Marc Khoury and Dylan Hadfield-Menell.
\newblock On the geometry of adversarial examples.
\newblock arXiv:1811.00525v1 [cs.LG], 2018.

\bibitem[Kos and Song(2017)]{kos:2017}
Jernej Kos and Dawn Song.
\newblock Delving into adversarial attacks on deep policies.
\newblock arXiv:1705.06452v1 [stat.ML], 2017.

\bibitem[Lanctot et~al.(2017)Lanctot, Zambaldi, Gruslys, Lazaridou, Tuyls,
  Perolat, Silver, and Graepel]{lanctot:2017}
Marc Lanctot, Vinicius Zambaldi, Audrunas Gruslys, Angeliki Lazaridou, Karl
  Tuyls, Julien Perolat, David Silver, and Thore Graepel.
\newblock A unified game-theoretic approach to multiagent reinforcement
  learning.
\newblock In \emph{Advances in Neural Information Processing Systems
  (NeurIPS)}, pages 4190--4203, 2017.

\bibitem[Lewis et~al.(2017)Lewis, Yarats, Dauphin, Parikh, and
  Batra]{lewis:2017}
Mike Lewis, Denis Yarats, Yann Dauphin, Devi Parikh, and Dhruv Batra.
\newblock Deal or no deal? {End-to-end} learning of negotiation dialogues.
\newblock In \emph{Proceedings of the Conference on Empirical Methods in
  Natural Language Processing (EMNLP)}, 2017.

\bibitem[Lin et~al.(2017)Lin, Hong, Liao, Shih, Liu, and Sun]{lin:2017}
Yen-Chen Lin, Zhang-Wei Hong, Yuan-Hong Liao, Meng-Li Shih, Ming-Yu Liu, and
  Min Sun.
\newblock Tactics of adversarial attack on deep reinforcement learning agents.
\newblock In \emph{Proceedings of the International Joint Conference on
  Artificial Intelligence (IJCAI)}, pages 3756--3762, 2017.

\bibitem[Maaten and Hinton(2008)]{maaten:2008}
Laurens van~der Maaten and Geoffrey Hinton.
\newblock Visualizing data using {t-SNE}.
\newblock \emph{Journal of Machine Learning Research (JMLR)}, 9\penalty0
  (Nov):\penalty0 2579--2605, 2008.

\bibitem[Mandlekar et~al.(2017)Mandlekar, Zhu, Garg, Fei-Fei, and
  Savarese]{mandlekar:2017}
Ajay Mandlekar, Yuke Zhu, Animesh Garg, Li~Fei-Fei, and Silvio Savarese.
\newblock Adversarially robust policy learning: Active construction of
  physically-plausible perturbations.
\newblock In \emph{Proceedings of the IEEE/RSJ International Conference on
  Intelligent Robots and Systems (IROS)}, pages 3932--3939, 2017.

\bibitem[Noonan(2017)]{noonan:2017}
Laura Noonan.
\newblock {JPMorgan} develops robot to execute trades.
\newblock \emph{Financial Times}, July 2017.

\bibitem[OpenAI(2018)]{openai:2018}
OpenAI.
\newblock {OpenAI Five}.
\newblock \url{https://blog.openai.com/openai-five/}, 2018.

\bibitem[Pattanaik et~al.(2018)Pattanaik, Tang, Liu, Bommannan, and
  Chowdhary]{pattanaik:2018}
Anay Pattanaik, Zhenyi Tang, Shuijing Liu, Gautham Bommannan, and Girish
  Chowdhary.
\newblock Robust deep reinforcement learning with adversarial attacks.
\newblock In \emph{Proceedings of the International Conference on Autonomous
  Agents and MultiAgent System (AAMAS)}, pages 2040--2042, 2018.

\bibitem[Pinto et~al.(2017)Pinto, Davidson, Sukthankar, and Gupta]{pinto:2017}
Lerrel Pinto, James Davidson, Rahul Sukthankar, and Abhinav Gupta.
\newblock Robust adversarial reinforcement learning.
\newblock In \emph{Proceedings of the International Conference on Machine
  Learning (ICML)}, volume~70, pages 2817--2826, 2017.

\bibitem[Ravanbakhsh and Sankaranarayanan(2016)]{ravanbakhsh:2016}
Hadi Ravanbakhsh and Sriram Sankaranarayanan.
\newblock Robust controller synthesis of switched systems using counterexample
  guided framework.
\newblock In \emph{Proceedings of the International Conference on Embedded
  Software (EMSOFT)}, pages 8:1--8:10, 2016.

\bibitem[Schulman et~al.(2017)Schulman, Wolski, Dhariwal, Radford, and
  Klimov]{schulman:2017}
John Schulman, Filip Wolski, Prafulla Dhariwal, Alec Radford, and Oleg Klimov.
\newblock Proximal policy optimization algorithms.
\newblock arXiv:1707.06347v2 [cs.LG], 2017.

\bibitem[Shafahi et~al.(2019)Shafahi, Huang, Studer, Feizi, and
  Goldstein]{shafahi:2019}
Ali Shafahi, W.~Ronny Huang, Christoph Studer, Soheil Feizi, and Tom Goldstein.
\newblock Are adversarial examples inevitable?
\newblock In \emph{Proceedings of the International Conference on Learning
  Representations (ICLR)}, 2019.

\bibitem[Shapley(1953)]{shapley:1953}
Lloyd~S. Shapley.
\newblock Stochastic games.
\newblock \emph{Proceedings of the National Academy of Sciences}, 39\penalty0
  (10):\penalty0 1095--1100, 1953.

\bibitem[Silver et~al.(2018)Silver, Hubert, Schrittwieser, Antonoglou, Lai,
  Guez, Lanctot, Sifre, Kumaran, Graepel, Lillicrap, Simonyan, and
  Hassabis]{silver:2018}
David Silver, Thomas Hubert, Julian Schrittwieser, Ioannis Antonoglou, Matthew
  Lai, Arthur Guez, Marc Lanctot, Laurent Sifre, Dharshan Kumaran, Thore
  Graepel, Timothy Lillicrap, Karen Simonyan, and Demis Hassabis.
\newblock A general reinforcement learning algorithm that masters chess, shogi,
  and {Go} through self-play.
\newblock \emph{Science}, 362\penalty0 (6419):\penalty0 1140--1144, 2018.

\bibitem[Szegedy et~al.(2014)Szegedy, Zaremba, Sutskever, Bruna, Erhan,
  Goodfellow, and Fergus]{szegedy:2014}
Christian Szegedy, Wojciech Zaremba, Ilya Sutskever, Joan Bruna, Dumitru Erhan,
  Ian~J. Goodfellow, and Rob Fergus.
\newblock Intriguing properties of neural networks.
\newblock In \emph{Proceedings of the International Conference on Learning
  Representations (ICLR)}, 2014.

\bibitem[Tamar et~al.(2014)Tamar, Mannor, and Xu]{tamar:2014}
Aviv Tamar, Shie Mannor, and Huan Xu.
\newblock Scaling up robust {MDPs} using function approximation.
\newblock In \emph{Proceedings of the International Conference on Machine
  Learning (ICML)}, pages II--181--II--189, 2014.

\bibitem[Uesato et~al.(2018)Uesato, O'Donoghue, Kohli, and van~den
  Oord]{uesato:2018}
Jonathan Uesato, Brendan O'Donoghue, Pushmeet Kohli, and Aaron van~den Oord.
\newblock Adversarial risk and the dangers of evaluating against weak attacks.
\newblock In \emph{Proceedings of the International Conference on Machine
  Learning (ICML)}, volume~80, pages 5025--5034, 2018.

\bibitem[Xie et~al.(2019)Xie, Wu, van~der Maaten, Yuille, and He]{xie:2019}
Cihang Xie, Yuxin Wu, Laurens van~der Maaten, Alan Yuille, and Kaiming He.
\newblock Feature denoising for improving adversarial robustness.
\newblock In \emph{Proceedings of the IEEE Conference on Computer Vision and
  Pattern Recognition (CVPR)}, 2019.

\end{thebibliography}
